\documentclass[10pt,twocolumn,letterpaper]{article}

\usepackage{cvpr}
\usepackage{times}
\usepackage{epsfig}
\usepackage{graphicx}
\usepackage{amsmath}
\usepackage{amssymb}

\usepackage{adjustbox}
\usepackage{multirow}

\usepackage[pagebackref=true,breaklinks=true,letterpaper=true,colorlinks,bookmarks=false]{hyperref}

\cvprfinalcopy 

\def\cvprPaperID{19} 

\ifcvprfinal\fi
\begin{document}

\title{Global Sum Pooling: A Generalization Trick for \\Object Counting with Small Datasets of Large Images}

\author{
  Shubhra Aich~ and~ Ian Stavness\\
  Department of Computer Science\\
  University of Saskatchewan, Canada\\
  \texttt{\{s.aich,ian.stavness\}@usask.ca} \\
}

\maketitle

\begin{abstract}
In this paper, we explore the problem of training one-look regression models for counting objects in datasets comprising a small number of high-resolution, variable-shaped images. We illustrate that conventional global average pooling (GAP) based models are unreliable due to the patchwise cancellation of true overestimates and underestimates for patchwise inference. To overcome this limitation and reduce overfitting caused by the training on full-resolution images, we propose to employ global sum pooling (GSP) instead of GAP or fully connected (FC) layers at the backend of a convolutional network. Although computationally equivalent to GAP, we show through comprehensive experimentation that GSP allows convolutional networks to learn the counting task as a simple linear mapping problem generalized over the input shape and the number of objects present. This generalization capability allows GSP to avoid both patchwise cancellation and overfitting by training on small patches and inference on full-resolution images as a whole. We evaluate our approach on four different aerial image datasets -- two car counting datasets (CARPK and COWC), one crowd counting dataset (ShanghaiTech; parts A and B) and one new challenging dataset for wheat spike counting.
Our GSP models improve upon the state-of-the-art approaches on all four datasets with a simple architecture. Also, GSP architectures trained with smaller-sized image patches exhibit better localization property due to their focus on learning from smaller regions while training.
\end{abstract}

\section{Introduction}

Increasingly complex and large deep learning architectures are being devised to tackle challenging computer vision problems, such as object detection and instance segmentation with hundreds of object classes~\cite{openimages,coco-objects,coco-stuff}. However, it is becoming common to deploy highly complex state-of-the-art architectures to solve substantially simpler tasks. Object counting is one such task: counting cars on a freeway or in a parking lot, counting people in a crowd, and counting plants or trees from aerial images. While it is possible to apply very powerful instance segmentation~\cite{mask-rcnn} or object detection~\cite{faster-rcnn} approaches to counting problems, these architectures require detailed (and time-consuming and tedious to collect) annotations, such as instance segmentation masks or bounding boxes. However, object counting is amenable to weaker labels, such as dot annotations (one dot per instance) or a scalar count per image. Devising simpler deep learning models for less complex computer vision tasks has the benefit of less costly ground-truth labeling, smaller sized networks, more efficient training, and faster inference.

One-look regression models are a class of deep neural network that are well matched to the comparatively simpler problem of object counting. These models use a convolutional front-end combined with fully-connected (FC) or global average pooling (GAP) layers that end in a single unit to generate a scalar count of the number of object instances present in the image \cite{aich-cvppp,dpp,deepwheat,best-cvppp}. Other variants of this counting network use a final classification layer, where the number of the output units are slightly more than the maximum number of possible object instances in the input \cite{cowc}. This requires that the maximum number of object instances are known \textit{a priori}, which may be difficult when the number of objects varies with the size of the input. Therefore, in this paper, we focus only on the single output unit models for object counting.

Counting datasets have two common characteristics that complicate the training of one-look models. First, the training set typically consists of a few very high-resolution images. Despite the computational complexity, it might be possible to train on full-sized images as a whole, but there is a high probability of overfitting by blindly memorizing the scalar counts because of the small number of training samples available.
Second, images with variable resolution in a single dataset are prevalent because they are often stitched or cropped to a particular region of interest. Many architectures require a pre-defined size for training/test images, and warping images to that pre-defined resolution could make small object instances almost disappear or large object instances become unrealistically large. In this way, downsampling and/or warping of training images can make the counting problem harder by increasing the ratio of resolution of larger to smaller instances present in the image dataset.

A common solution to overcome the challenge of high-resolution, variable-sized images is to use smaller sized, randomly cropped ``patches'' from the high-resolution raw training images to train the network. Dot annotations can be counted to create an object count per patch, but because the full extent of the object is not annotated, it is not possible to generate patches without partially cutting the objects at the edge of the patch. This type of label noise may be acceptable during training, but at test time, when a total count is required for the high-resolution image, tiled patches would need to be applied to the network with global average pooling (GAP) layers in the backend and the counts per tile summed. We demonstrate later in the experiments section that using GAP with fixed-resolution, smaller patches incorporates both per-patch underestimates and overestimates. Aggregating the counts of all the patches in a single image randomly cancels-out a large number of such overestimates and underestimates; thus giving an apparent impression of a reasonably accurate measure for each image, but obscuring substantial per-patch inference errors. The extent of patchwise cancellation of positive errors by negative errors is random and depends on the pattern of the collocation of the object instances in the images. Also, for rectangular patches, per-patch overestimates and underestimates increase for images with objects not oriented in a spatially vertical or horizontal fashion. We hypothesize that all these shortcomings make the use of patchwise inference with GAP unreliable for counting objects with small datasets of high-resolution images.
Previous work has attempted to resolve these patch-wise inference errors empirically, by optimizing the stride of tiled patches based on validation set performance~\cite{cowc}. However, this does not address the fundamental limitation of partial object instances; resulting in unavoidable per-patch counting errors, which are then propagated to the estimate of the full image count.

\begin{figure*}[t]
\centering
\includegraphics[scale=0.38,angle=0]{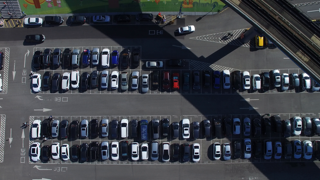}
\includegraphics[scale=0.38,angle=0]{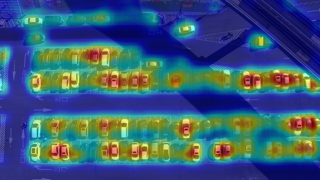}
\includegraphics[scale=0.38,angle=0]{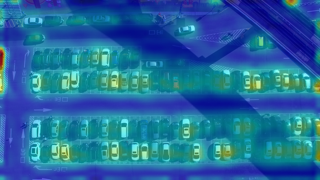}
\includegraphics[scale=0.38,angle=0]{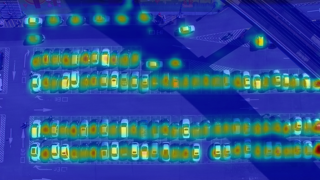}
\caption{Sample image for car counting \cite{lpn-carpk} along with superimposed activation heatmaps for different one-look regression models, from left-to-right: original image, the baseline GAP model, our GSP model trained with full-resolution images, and GSP trained with 224~$\times$~224 randomly cropped patches.
}
\label{fig:detector}
\end{figure*}

Considering the complications of datasets with a small number of high-resolution variable-sized images, an ideal solution would be a particular kind of model that can be trained with small-sized random patches (to reduce the risk of overfitting or memorization) and then generalize its performance over arbitrarily large resolution test samples.
In this paper, we devise such a model using a set of traditional convolutional and pooling layers in the front-end and replacing the fully connected (FC) layers or global average pooling (GAP) layer with the new global sum pooling (GSP) operation. We show that the use of this GSP layer allows the network to train on image patches and infer accurate object counts on full sized images. Although from a computational perspective, the summation operation in GSP is very similar to the averaging operation in GAP, GSP exhibits the non-trivial property of generalization for counting objects over variable input shapes, which GAP does not. To the best of our knowledge, this is the first work introducing the GSP operation as a replacement of GAP or FC layers. We evaluated GSP models on four different datasets --- two for counting cars, one for crowd counting, and one for counting wheat spikes. Our experimental results demonstrate that GSP helps to generate more localized activations on object regions (Figure~\ref{fig:detector}) and achieve better generalization performance which is consistent with our hypothesis.
\\

\begin{figure}[]
\centering
\includegraphics[scale=0.160]{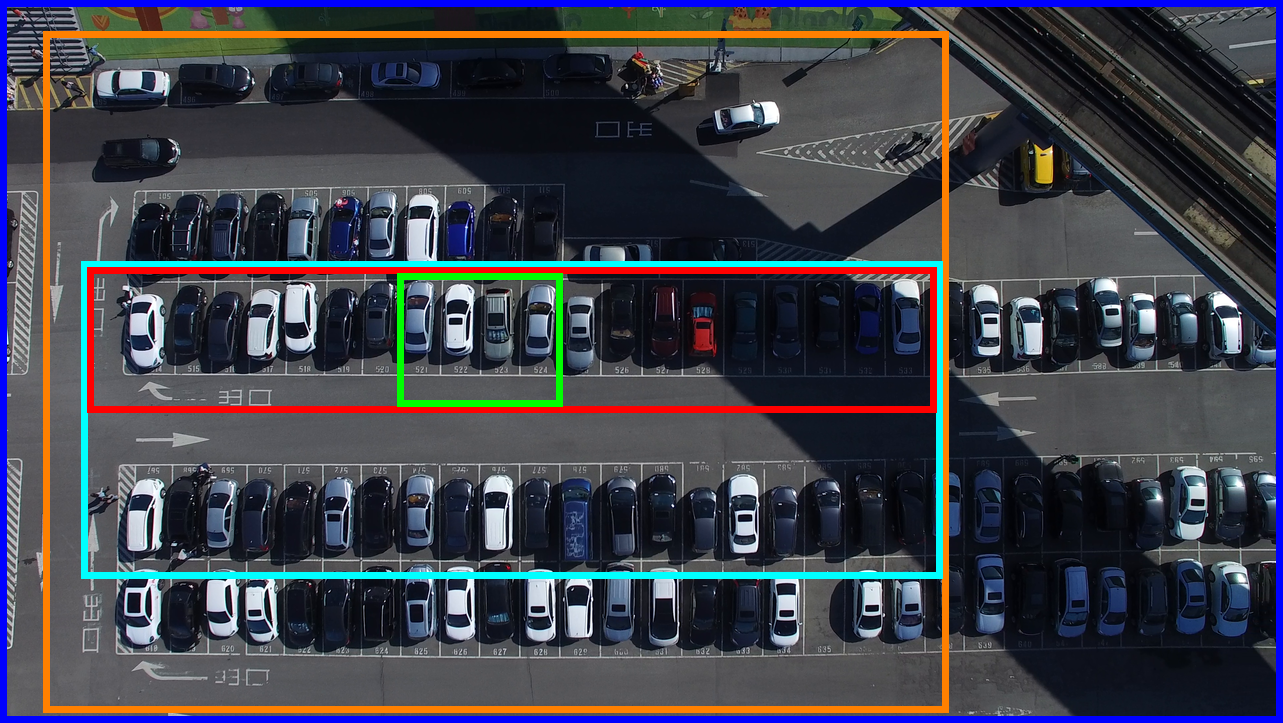} \\
\includegraphics[scale=0.58]{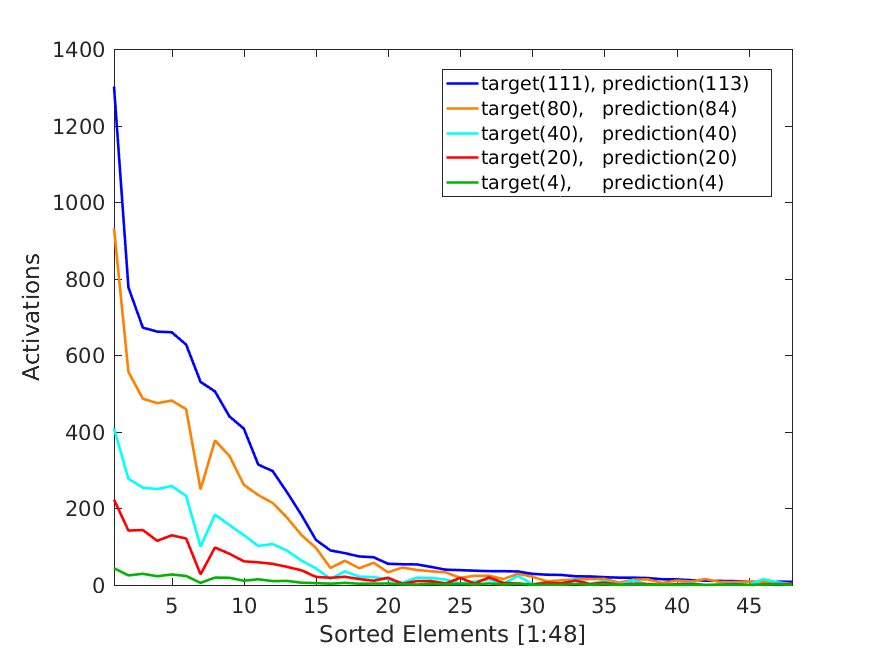}
\caption{(Left) Sample image with multiple cropping shown using bounding boxes with different colors. (Right) Activations of the first 48 elements sorted in descending order incurred by these cropped samples after GSP operation shown using the corresponding colors of the bounding boxes in the left. For consistency, sorting indices of the full-resolution input are used to sort others. The plot of the values demonstrates the fact of learning a linear mapping of the object counts by our GSP-CNN model regardless of input shape. }
\label{fig:lin_scale}
\end{figure}

\noindent Our paper makes the following contributions:

\begin{itemize}

\item We describe the limitations of existing architectural designs for object counting on datasets with fewer, high-resolution training samples. With extensive experimentation, we demonstrate the patchwise cancellation effect on per-patch overestimates and underestimates of using global average pooling (GAP) with tiled patches. We argue that such randomization makes GAP a fragile candidate for counting architectures.

\item We propose global sum pooling (GSP) as an alternative to GAP or FC layers as a remedy to the problems regarding variable resolution images, and overfitting on small datasets of large images. We demonstrate that GSP models can be trained on smaller random patches and used for inference on full resolution images. This is because the GSP models learn a mapping linear to the number of objects regardless of input resolution, which is impossible with GAP or FC layer models.

\item We benchmark GSP on four heterogeneous (two car parks, one crowd counting, and one wheat spike counting) datasets. Our GSP model beats state-of-the-art approaches with much better saliency mapping on all these datasets. Results were obtained with a simple convolutional front-end which demonstrates the simplicity and efficacy of GSP for object counting.
\end{itemize}

\section{Related Work}

Much of the recent literature on object counting is based on estimating different kinds of activation maps because these approaches are applicable to datasets with high-resolution images.
Lempitsky and Zisserman \cite{zisserman-nips-count} incorporate the idea of per-pixel density map estimation followed by regression for object counting. This regression approach is further enhanced by \cite{interactive-count} by adding an interactive user interface. Fiaschi et al. \cite{fiaschi2012} employ random forest to regress the density map and object count. Fully convolutional network \cite{xie2016} is also used for contextual density map estimation irrespective of the input shape. Proximity map, which is the proximity to the nearest cell center, is also estimated in \cite{xie2015} as an alternative to traditional density map approximation. Another variant of density map is proposed in the Count-ception paper \cite{countception}, where the authors use fully convolutional network \cite{fcn} to regress the count map followed by scalar count retrieval adjusting the redundant coverage proportional to the kernel size. Wheat spike images have been previously investigated for controlled imaging environments using density maps~\cite{average-pridmore}.

There also exists an extensive body of work on crowd counting \cite{survey-crowd}. Here, we review some of the recent CNN based approaches. Wang et al. \cite{acm-multimedia-2015} employed a one-look CNN model first on dense crowd counting. Zhang et al. \cite{cross-scene-sjtu} developed the \emph{cross-scene} crowd counting approach. They use alternative optimization criteria for counting and density map estimation. Also, instead of single Gaussian kernels to generate a ground truth density map, they use multiple kernels along with the idea of perspective normalization. Cross-scene adaptation is done by finetuning the network with training samples similar to test scenes. Similar to the gradient boosting machines \cite{gradient-boosting-machine}, Walach and Wolf \cite{crowd-cnn-boosting} iteratively add additional computational blocks in their architecture to train on the residual error of the previous block, which they call layered boosting. Shang et al. \cite{shang-local-global} use an LSTM \cite{lstm} decoder on GoogLeNet \cite{inception-v1} features to extract a patchwise local count and generate a global count from them using FC layers. CrowdNet \cite{crowdnet} uses the combination of shallow and deep networks to acquire multi-scale information in density map approximation for crowd counting. Another approach \cite{multi-column-crowd} for multi-scale context aggregation for density map estimation use multi-column networks with different kernel sizes. Hydra CNN \cite{hydra-cnn} employ three convolutional heads to process image pyramids and combine their outputs with additional FC layers to approximate the density map at a lower resolution. Switching CNN architecture \cite{switching-cnn} proposes a switching module to decide among different sub-networks to process images with different properties.

Most of the approaches described above attempt to approximate a final activation map under different names, i.e. density map, count map, and proximity map, which is then post-processed to obtain the count information; thus resulting in a multi-stage pipeline. In this regard, one-look models are simpler and faster than these map estimation approaches.
The main idea behind using one-look regression models~\cite{aich-cvppp,deepwheat,acm-multimedia-2015,dpp,best-cvppp,cowc} for object counting is to utilize weaker ground truth information like dot annotations, in contrast to more sophisticated models for object detection \cite{faster-rcnn,yolo} or instance-level segmentation \cite{mask-rcnn,ris,instance-seg-toronto} that require stronger and more tedious to collect ground truth labels. The domain knowledge of spatial collocation of cars in the car parks is exploited in the layout proposal network \cite{lpn-carpk} to detect and count cars. The COWC dataset paper \cite{cowc} uses multiple variants of the hybrid of residual \cite{resnet} and Inception \cite{inception-v3} architectures, called ResCeption, as the one-look model for counting cars patchwise. During inference, the authors determine the stride based on the validation set. This kind of hybrid models are also used in \cite{deepwheat} to estimate plant characteristics from images. However, recent work on heatmap regulation (HR) \cite{heatmap-regulation} describes the philosophical limitation of using one-look models and tries to improve its performance by regulating the final activation map with a Gaussian approximation of the ground-truth activation map. In this paper, using GSP and training with smaller samples, we obtain similar final activation maps to HR without using any extra supervision channel in our model. 


\section{Our Approach}

To overcome the generalization challenges for object counting from a small number of high-resolution, variable sized images, and to avoid the problem of partial object counting while cropping random patches, we propose an architecture that can learn to count from images regardless of their shape.
Architectures with final FC layers pose strict requirements about the input images' shape, whereas architectures that combine CNN with additional nonlinear and normalization layers are more flexible.
We take inspiration from recent image classification architectures \cite{nin,resnet,inception-v3,densenet} that replace FC layers with a simple GAP layer. Using GAP greatly reduces the number of parameters (to help reduce overfitting), emphasizes the convolutional front-end of the models, permits training and testing on variable size input images, and provides intuitive visualizations of the activation maps~\cite{cam-mit}.
%
For an object counting task, however, the averaging operation of GAP lacks the ability to generalize over variable resolution input images.

The difference between GAP and GSP for object counting can be illustrated by a hypothetical example. For simplicity of illustration, here we consider an ideal environment where the resolution of object instances falls within a fixed range over a dataset, but this is not a requirement for the GSP approach. Also, without loss of generality, we assume that objects are uniformly distributed, which means that the number of objects within an image is expected to scale with the image resolution. For example, if a $W\times W$ region contains $\mathcal{C}$ objects, then a $mW\times mW$ region would be expected to contain $m^2W$ objects ($m$ is a scalar). If we train a network containing a stack of convolution layers followed by a GAP layer on $W\times W$ samples, our models will learn to generate the expected count of $\mathcal{C}$ with an equivalent vector representation as the output of GAP. During inference, with a $mW\times mW$ image, the last convolution layer will generate $m^2$ adjacent, spatial feature responses, each representing the expected count of $\mathcal{C}$. This convolutional representation is appropriate to predict an expected count of $m^2\mathcal{C}$. However, the GAP layer will average over all the $m^2$ spatial sub-regions and obtain an equivalent representation of $\mathcal{C}$. Hence, the averaging operation is not suitable for modeling the proportional scaling of the number of objects with the size of input.

Another option might be to divide the variable resolution input images into fixed size, adjacent, and non-overlapping patches during inference and then sum up the count over all of the filled patches to retrieve the final count. Computationally, such tiling preserves the efficiency of inference in a single pass. Inference on overlapping patches with density estimation is also possible~\cite{countception}, but quadratically increases the computational complexity. Any patch-wise inference scheme, however, has a significant limitation that is unavoidable from the modeling perspective. During inference, the network produces both overestimates and underestimates over all the patches extracted from a single image. For example, for a single image, the amount of overestimates and underestimates are $\mathcal{E_O}$ and $\mathcal{E_U}$, respectively. Although the actual difference between ground truth count and prediction is $\mathcal{E_O} + \mathcal{E_U}$, by summing up the patch counts, we get an apparent error of $|\mathcal{E_O} - \mathcal{E_U}|$. Thus, the measured difference on the whole image, in this case, depends on the difference between overestimate $\mathcal{E_O}$ and underestimate $\mathcal{E_U}$, not on their absolute value. For example, we will get a very low error even if $\mathcal{E_O}$ and $\mathcal{E_U}$ are quite high but are almost equal. Thus, when aggregating the patch count, overestimate and underestimate get nullified by each other randomly on which the model has no control. We call this effect ``patchwise cancellation". The amount of such patchwise cancellation depends on various properties of the dataset, such as the density and types of the objects, their collocation patterns, their comparative resolution in the image, and so on. In the experiments section, we show that although for GAP models used on adjacent patches for inference, the difference between ground truth and summed up prediction is sometimes reasonably small, the actual overestimate and underestimate are pretty high, validating our explanation of patchwise cancellation. Therefore, GAP with adjacent patches is not a reliable solution for inference on variable-sized, high-resolution images.

\begin{table*}[]
\centering
\caption{Statistics of the datasets used for evaluation}
\label{tab:stat_dataset}
\begin{adjustbox}{width=0.75\textwidth,center}
\begin{tabular}{lcccc}
\hline
\multicolumn{1}{c}{Dataset} & \begin{tabular}[c]{@{}c@{}}\#Images\\ (Train, Test)\end{tabular} & Resolution & \begin{tabular}[c]{@{}c@{}}Total Count\\ (Train, Test)\end{tabular} & \begin{tabular}[c]{@{}c@{}}Range of Count\\ (Train, Test)\end{tabular} \\ \hline
CARPK & (989, 459) & (720~$\times$~1280) & (42274, 47500) & ({[}1, 87{]}, {[}2, 188{]}) \\ \hline
ShanghaiTech-A & (300, 182) & (200$\times$300) -- (1024$\times$992) & (162413, 78862) & ({[}33, 3138{]}, {[}66, 2256{]}) \\ \hline
ShanghaiTech-B & (400, 316) & (768~$\times$~1024) & (49151, 39121) & ({[}12, 576{]}, {[}9, 539{]}) \\ \hline
COWC & (32, 20) & $\sim$(18k~$\times$~18k) -- (2k~$\times$~2k) & (37890, 3456) & ({[}45, 13086{]}, {[}10, 881{]}) \\ \hline
Wheat-Spike & (10, 10) & $\sim$(1000, 3000) & (10112, 9989) & ({[}796, 1287{]}, {[}749, 1205{]}) \\ \hline
\end{tabular}
\end{adjustbox}
\end{table*}

Instead of average-pooling the final feature maps, we propose a summation or mere aggregation of the input over the spatial locations only. From the previous example, this aggregation of $m^2$ similar sub-regions, each with a count of $\mathcal{C}$, would produce the desired expected value of $m^2\mathcal{C}$. Following the nomenclature of GAP, we call this operation global sum pooling (GSP). Although GAP and GSP are computationally similar operations, conceptually GSP provides the ability to use CNN architectures for generalized training and inference on variable shaped inputs in a simple and elegant way. Moreover, due to the single pass inference regardless of input resolution, GSP does not suffer from patchwise cancellation.

\textbf{Linear mapping:}
Learning to count regardless of the input image shape necessarily means that the convolutional front-end of the network should learn a linear mapping task, where the output vector of GSP will scale proportionally with the number of objects present in the input image. Figure \ref{fig:lin_scale} shows a sample 720~$\times$~1280 image from the CARPK \cite{lpn-carpk} aerial car counting dataset. On the right of Figure \ref{fig:lin_scale}, we plot the largest 48 activations of the 512-vector output of the GSP layer of our model described later, for different-sized sub-regions of the same sample image. Here, the elements are sorted in descending order for the full resolution image, and the same ordering is used for the activations of the sub-regions. The model producing these activations was trained on 224~$\times$~224 randomly cropped samples. From this figure, it is evident that our model is able to learn a linear mapping function from the image space to the high-dimensional feature space, where the final count is a simple linear regression or combination of the extracted feature values.


\textbf{Weak instance detector and region classifier:}
An advantage of training on small input sizes is that it guides the network to behave like a weak object instance detector even though we only provide weak labels (a scalar count per image region).
%
Training on sub-regions of a large input image helps the network to better disambiguate the true object regions from the object-like background sub-regions, resulting in improved performance. For example, when training the network with full images, all of which have a non-zero object count, the network never faces a complete background sample from which it can extract background information similar to any binary region classification problem. On the other hand, when we train with small randomly-cropped regions of the input image many background-only samples are fed to the network, instantiating a more rigorous learning paradigm even with weak count labels.
Class-activation map (CAM) \cite{cam-mit} visualizations illustrate that the GSP model trained with small sub-regions better captures localization information (Figure \ref{fig:detector}). Training the GAP or GSP models with full-resolution images results in a less uniform distribution of activation among object regions and less localized activations inside object regions as compared to the GSP model trained with smaller patches.

\textbf{Architecture:}
We attach a GSP layer after the convolutional front-end of VGG16 \cite{vgg} model pretrained on ImageNet \cite{imagenet}. GSP produces a 512-dimensional vector, which is converted to a scalar count by a linear layer. We faced no problems with the potential numerical instability caused by large, unnormalized values after spatial summation, even when training the GSP models with full resolution images.

\section{Experiments}

\textbf{Datasets:} We evaluate object counting with GSP on four datasets: CARPK \cite{lpn-carpk} (overhead view of different car parks), ShanghaiTech \cite{multi-column-crowd} (crowd images collected from the web and streets of Shanghai), COWC \cite{cowc} (overhead view of cars in residential areas and highways), and a wheat spike (WS) dataset \cite{wheat-spike} (overhead view of mature wheat plants). CARPK and ShanghaiTech-B contain constant resolution images, whereas ShanghaiTech-A, COWC, and WS have large images with variable resolutions. All datasets have comparatively few training and test images. Statistics of these datasets are listed in Table \ref{tab:stat_dataset}.

\textbf{Metrics:} We adopt the evaluation metrics from MCNN \cite{multi-column-crowd} and COWC \cite{cowc} papers along with one additional metric: the percentage of MAE over expected ground truth, which we call the relative MAE (\%RMAE) (Equation \ref{eq:metrics}).

\begin{equation}
\begin{cases}
\textit{Mean Absolute Error (MAE)} = \frac{\sum_{i}|\hat{y_i}-y_i|}{N} \\
\textit{\%MAE} = \frac{\sum_{i}|\hat{y_i}-y_i|}{Ny_i} \times 100 \\
\textit{Relative MAE (\%RMAE)} = \frac{MAE \times N}{\sum_{i}y_i} \times 100 \\
\textit{Root-Mean-Square Error (RMSE)} = \sqrt{\frac{\sum_{i}\left(\hat{y_i}-y_i\right)^{2}}{N} } \\
\textit{\%RMSE} = \sqrt{\frac{\sum_{i}\left(\hat{y_i}-y_i\right)^{2}}{Ny_i^2} } \times 100 \\
\end{cases}
\label{eq:metrics}
\end{equation}

\textbf{Models \& Training:} We train both GSP and GAP models on full-resolution images and on randomly cropped patches of resolutions $224$, $128$, $96$, and $64$. For GSP models, inference is done on full resolution images regardless of their shape and input size used at training, which is not possible for GAP models. For GAP models, we provide error metrics in two forms. First, we report errors over the cumulative patch counts that we denote by \textit{GAP-C (GAP-Cumulative)}. However, as described before, such error is a misinterpretation of the actual per patch error of the GAP models. Therefore, another error is estimated per tiled patch and all the per patch errors over the single image are summed up under the tag of \textit{GAP-PS (GAP-Patch-Summed)}.

In order to train on image patches, we compute a count per patch based on the number of central object regions within the patch. The CARPK dataset provides bounding boxes, which we shrink down to 25\% along each dimension and to define a central region for each car instance. The shrinking prevents object regions from overlapping and makes it so that we only count objects that are mostly inside the cropped patch. The ShanghaiTech, COWC, and Wheat-Spike datasets provide dot annotations appropriate to train our models.

\begin{figure}[t]
\centering
\includegraphics[scale=0.21,angle=90]{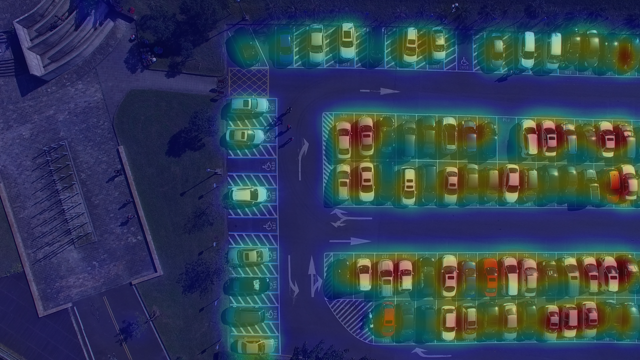}
\includegraphics[scale=0.21,angle=90]{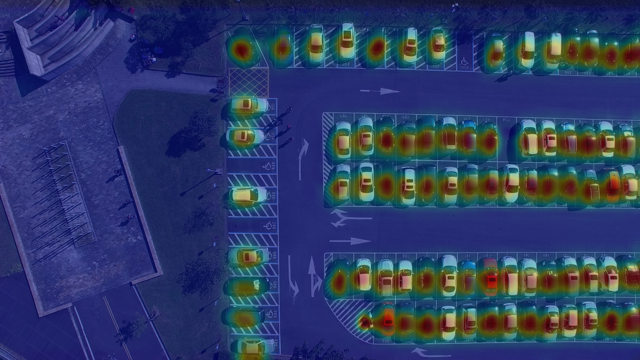}
\includegraphics[scale=0.21,angle=90]{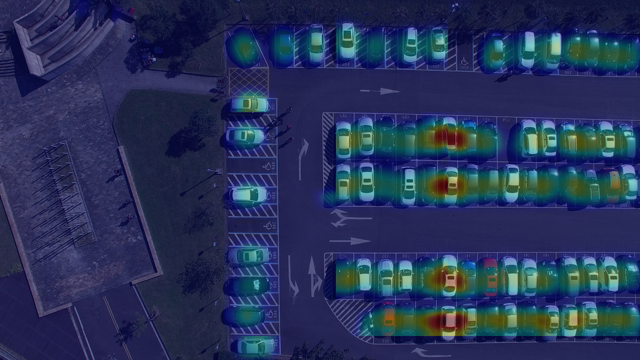}
\caption{Activation maps for CARPK generated by the GAP-Full (left), GSP-224 (middle), and GAP-224(right) models.
Activations are more uniformly distributed and more concentrated inside object regions for the GSP-224 model.}
\label{fig:carpk}
\end{figure}

\begin{table}[]
\centering
\caption{GSP-GAP comparison on CARPK dataset}
\label{tab:carpk-all}
\begin{adjustbox}{width=0.48\textwidth,center}
\begin{tabular}{c|l|ccccc}
\hline
\multicolumn{1}{l|}{Input} & Type & \multicolumn{1}{c}{MAE} & \multicolumn{1}{c}{RMSE} & \multicolumn{1}{c}{\%MAE} & \multicolumn{1}{c}{\%RMSE} & \%RMAE \\ \hline
\multirow{2}{*}{Full} & GAP & 19.61 & 21.65 & 23.78 & 42.87 & 18.95 \\ \cline{2-7}
 & GSP & 32.94 & 36.23 & 39.46 & 70.91 & 31.83 \\ \hline
\multirow{3}{*}{224} & GAP-C & 7.65 & 9.59 & 9.34 & 15.65 & 7.39 \\ \cline{2-7}
 & GAP-PS & 19.20 & 21.42 & 19.01 & 26.38 & 16.67 \\ \cline{2-7}
 & GSP & \textbf{5.46} & \textbf{8.09} & \textbf{12.21} & \textbf{44.36} & \textbf{5.28} \\ \hline
\multirow{3}{*}{128} & GAP-C & 8.66 & 11.30 & 11.90 & 26.64 & 8.37 \\ \cline{2-7}
 & GAP-PS & 22.14 & 25.94 & 21.43 & 34.05 & 17.88 \\ \cline{2-7}
 & GSP & 6.70 & 10.21 & 8.74 & 19.15 & 6.48 \\ \hline
\multirow{3}{*}{96} & GAP-C & 10.72 & 13.63 & 17.23 & 49.35 & 10.36 \\ \cline{2-7}
 & GAP-PS & 44.41 & 48.49 & 39.23 & 54.88 & 32.89 \\ \cline{2-7}
 & GSP & 10.63 & 11.37 & 22.27 & 62.87 & 10.27 \\ \hline
\multirow{3}{*}{64} & GAP-C & 23.20 & 27.78 & 42.16 & 115.61 & 22.42 \\ \cline{2-7}
 & GAP-PS & 52.81 & 57.24 & 52.51 & 110.44 & 34.98 \\ \cline{2-7}
 & GSP & 32.09 & 36.02 & 31.64 & 34.39 & 31.01 \\ \hline
\end{tabular}
\end{adjustbox}
\end{table}

\begin{table}[]
\centering
\caption{Results on CARPK dataset}
\label{tab:carpk}
\begin{adjustbox}{width=0.35\textwidth,center}
\begin{tabular}{l|cc}
\hline
Method & MAE & RMSE \\ \hline
YOLO \cite{yolo,lpn-carpk} & 48.89 & 57.55 \\ \hline 
Faster R-CNN \cite{faster-rcnn,lpn-carpk} & 47.45 & 57.39 \\ \hline 
One-Look Regression \cite{cowc,lpn-carpk} & 59.46 & 66.84 \\ \hline 
LPN \cite{lpn-carpk} & 13.72 & 21.77 \\ \hline 
HR \cite{heatmap-regulation} & 7.88 & 9.30 \\ \hline
Ours (GSP-224) & \textbf{5.46} & \textbf{8.09} \\ \hline
\end{tabular}
\end{adjustbox}
\end{table}

\textbf{CARPK dataset: }
For this car park dataset, we found that GSP models trained with 128~$\times$~128 and 224~$\times$~224 samples perform much better than the same model trained with smaller patches, like 64~$\times$~64 and 96~$\times$~96 (Table \ref{tab:carpk}). The reason for GAP-C showing apparently lower error is the patchwise cancellation of patchwise overestimate and underestimate of GAP models which is evident from the numerics of GAP-PS.

Figure \ref{fig:carpk} compares CAM heatmaps superimposed on original images for the baseline GAP-Full model and our best performing GSP-N model and GAP-N (N=224 for both). The activation maps of the GAP model are variable over the object regions, indicating that some of the objects are being highly emphasized than others, whereas the GSP-224 activations are more uniform, showing that all the instances are getting more or less equal attention from the network. Moreover, the GSP-224 activations better localized within object sub-regions than GAP model, which demonstrates that GSP-N models with small N work as a better object detector or binary region classifier than the baseline models.

We believe that the poor performance of GSP-N models for smaller N (64 and 96) is not a characteristic of the model itself. Instead, the poor performance can be attributed to the training procedure that we followed in this paper. As already stated, we shrink the bounding boxes for CARPK dataset to disambiguate the overlapping bounding boxes. However, such shrinking poses restrictions on using arbitrarily small sample sizes in training. If the patch size is close to the object resolution (the average resolution of the bounding boxes in the training set of CARPK dataset is about 40 pixels) and the objects are close together (which cars are in a parking lot), a patch is likely to include one complete object with several other instances partially cut at the edge of the patch. Because we disambiguate object counts by shrinking the boxes, depending on the relative orientation between the object and its encompassing box, and its portion inside the cropped patch, it might be taken into account for counting or not. Therefore, this aspect of our training paradigm is a bit randomized. For comparatively larger sample size, such as 128 and 224, we anticipate that this problem of random consideration of the partial objects in the border is less frequent than the smaller sized patches, such as 64 and 96. In this regard, the optimal sample size depends on the average resolution of the object instances and their relative placement in the images of a particular dataset.

\begin{table}[t]
\centering
\caption{GSP-GAP comparison on ShanghaiTech-A dataset}
\label{tab:a-shanghaitech-all}
\begin{adjustbox}{width=0.48\textwidth,center}
\begin{tabular}{c|l|ccccc}
\hline
\multicolumn{1}{l|}{Input} & Type & \multicolumn{1}{c}{MAE} & \multicolumn{1}{c}{RMSE} & \multicolumn{1}{c}{\%MAE} & \multicolumn{1}{c}{\%RMSE} & \%RMAE \\ \hline
\multirow{2}{*}{Full} & GAP & 143.13 & 199.79 & 43.21 & 64.84 & 33.09 \\ \cline{2-7}
 & GSP & 153.38 & 259.04 & 31.98 & 38.91 & 35.46 \\ \hline
\multirow{3}{*}{224} & GAP-C & 83.50 & 124.39 & 23.01 & 34.89 & 19.31 \\ \cline{2-7}
 & GAP-PS & 104.13 & 140.67 & 28.03 & 37.37 & 24.08 \\ \cline{2-7}
 & GSP & \textbf{70.69} & \textbf{103.58} & \textbf{19.66} & \textbf{29.37} & \textbf{16.34} \\ \hline
\multirow{3}{*}{128} & GAP-C & 83.75 & 124.79 & 23.28 & 34.27 & 19.36 \\ \cline{2-7}
 & GAP-PS & 113.31 & 148.58 & 30.63 & 38.50 & 26.20 \\ \cline{2-7}
 & GSP & 71.19 & 111.86 & 18.73 & 29.28 & 16.46 \\ \hline
\multirow{3}{*}{96} & GAP-C & 86.81 & 124.88 & 23.31 & 30.71 & 20.07 \\ \cline{2-7}
 & GAP-PS & 127.63 & 162.94 & 33.35 & 37.79 & 29.51 \\ \cline{2-7}
 & GSP & 78.25 & 116.69 & 19.87 & 27.21 & 18.09 \\ \hline
\multirow{3}{*}{64} & GAP-C & 100.69 & 140.53 & 26.87 & 33.80 & 23.28 \\ \cline{2-7}
 & GAP-PS & 160.38 & 197.79 & 42.16 & 45.99 & 37.08 \\ \cline{2-7}
 & GSP & 107.81 & 151.72 & 29.98 & 37.97 & 24.93 \\ \hline
\end{tabular}
\end{adjustbox}
\end{table}

\begin{table}[h!]
\centering
\caption{GSP-GAP comparison on ShanghaiTech-B dataset}
\label{tab:b-shanghaitech-all}
\begin{adjustbox}{width=0.48\textwidth,center}
\begin{tabular}{c|l|ccccc}
\hline
\multicolumn{1}{l|}{Input} & Type & \multicolumn{1}{c}{MAE} & \multicolumn{1}{c}{RMSE} & \multicolumn{1}{c}{\%MAE} & \multicolumn{1}{c}{\%RMSE} & \%RMAE \\ \hline
\multirow{2}{*}{Full} & GAP & 12.91 & 20.19 & 13.44 & 21.66 & 10.45 \\ \cline{2-7}
 & GSP & 12.26 & 19.49 & 12.01 & 19.56 & 9.92 \\ \hline
\multirow{3}{*}{224} & GAP-C & 9.70 & 16.03 & 7.69 & 10.16 & 7.84 \\ \cline{2-7}
 & GAP-PS & 18.44 & 24.04 & 15.87 & 17.38 & 14.91 \\ \cline{2-7}
 & GSP & 9.96 & 16.67 & 8.07 & 10.71 & 8.06 \\ \hline
\multirow{3}{*}{128} & GAP-C & 10.24 & 16.81 & 8.43 & 11.34 & 8.29 \\ \cline{2-7}
 & GAP-PS & 24.05 & 30.31 & 21.24 & 22.76 & 19.45 \\ \cline{2-7}
 & GSP & \textbf{9.13} & \textbf{15.94} & \textbf{7.05} & \textbf{9.24} & \textbf{7.39} \\ \hline
\multirow{3}{*}{96} & GAP-C & 11.16 & 17.60 & 9.31 & 11.93 & 9.03 \\ \cline{2-7}
 & GAP-PS & 28.73 & 34.74 & 25.49 & 26.76 & 23.24 \\ \cline{2-7}
 & GSP & 9.48 & 15.40 & 7.70 & 10.21 & 7.67 \\ \hline
\multirow{3}{*}{64} & GAP-C & 15.94 & 21.57 & 15.69 & 18.69 & 12.90 \\ \cline{2-7}
 & GAP-PS & 39.06 & 46.88 & 35.30 & 36.75 & 31.60 \\ \cline{2-7}
 & GSP & 14.35 & 22.95 & 12.44 & 15.85 & 11.61 \\ \hline
\end{tabular}
\end{adjustbox}
\end{table}

\begin{table}[]
\centering
\caption{Results on ShanghaiTech dataset}
\label{tab:shanghaitech}
\begin{adjustbox}{width=0.40\textwidth,center}
\begin{tabular}{l|cc|cc}
\hline
\multicolumn{1}{c|}{\multirow{2}{*}{Method}} & \multicolumn{2}{c|}{Part A} & \multicolumn{2}{c}{Part B} \\ \cline{2-5}
\multicolumn{1}{c|}{} & MAE & RMSE & MAE & RMSE \\ \hline
CS-CNN \cite{cross-scene-sjtu} & 181.8 & 277.7 & 32.0 & 49.8 \\ \hline
MCNN \cite{multi-column-crowd} & 110.2 & 173.2 & 26.4 & 41.3 \\ \hline
FCN \cite{fcn-marsden} & 126.5 & 173.5 & 23.76 & 33.12 \\ \hline
Cascaded MTL \cite{cascaded-mtl} & 101.3 & 152.4 & 20.0 & 31.1 \\ \hline
Switching-CNN \cite{switching-cnn} & 90.4 & 135.0 & 21.6 & 33.4 \\ \hline
Ours (GSP-224 and -128) & \textbf{70.7} & \textbf{103.6} & \textbf{9.1} & \textbf{15.9} \\ \hline
\end{tabular}
\end{adjustbox}
\end{table}

\begin{figure}[t]
\centering
\includegraphics[scale=0.11]{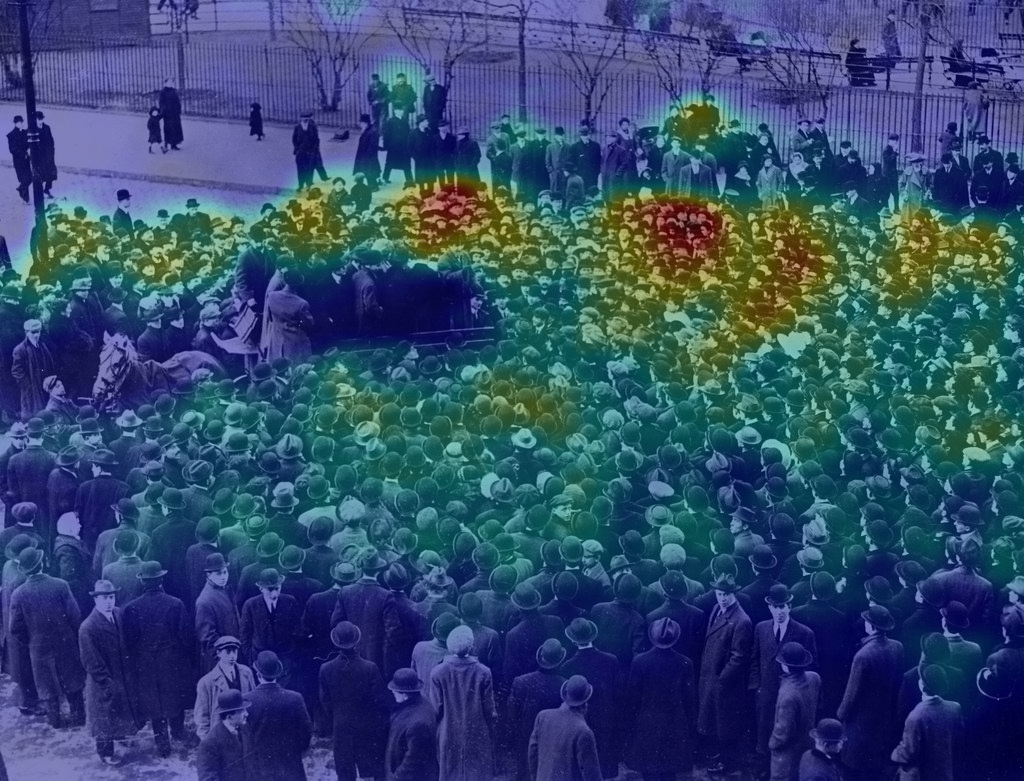}
\includegraphics[scale=0.11]{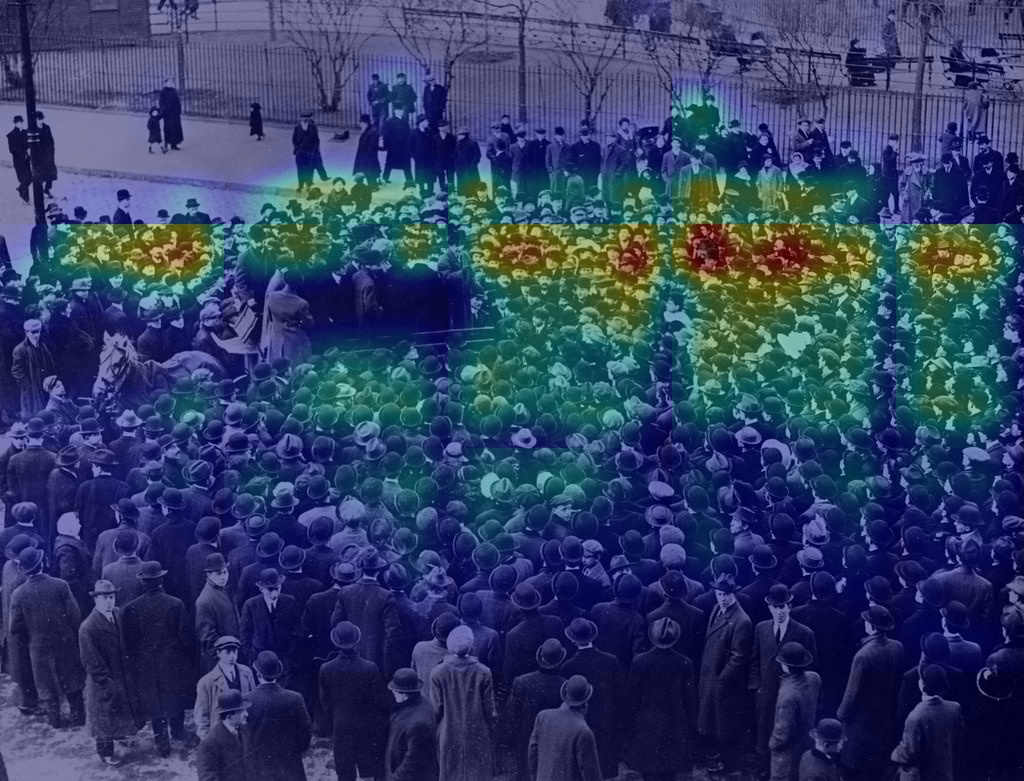} \\
\includegraphics[scale=0.11]{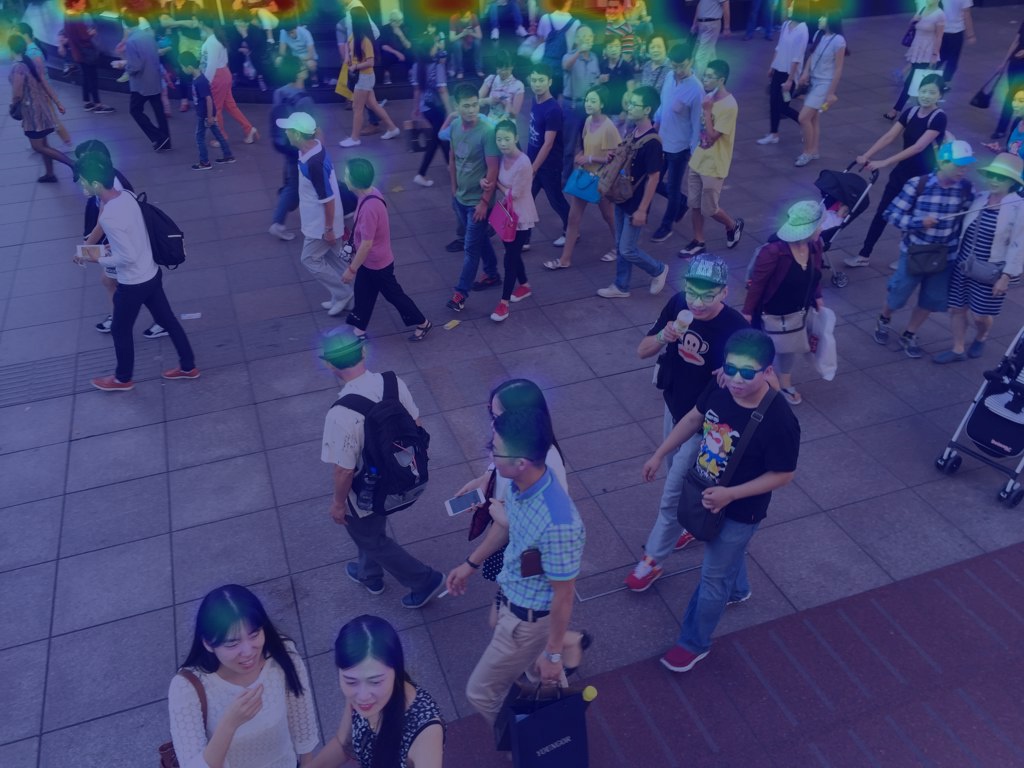}
\includegraphics[scale=0.11]{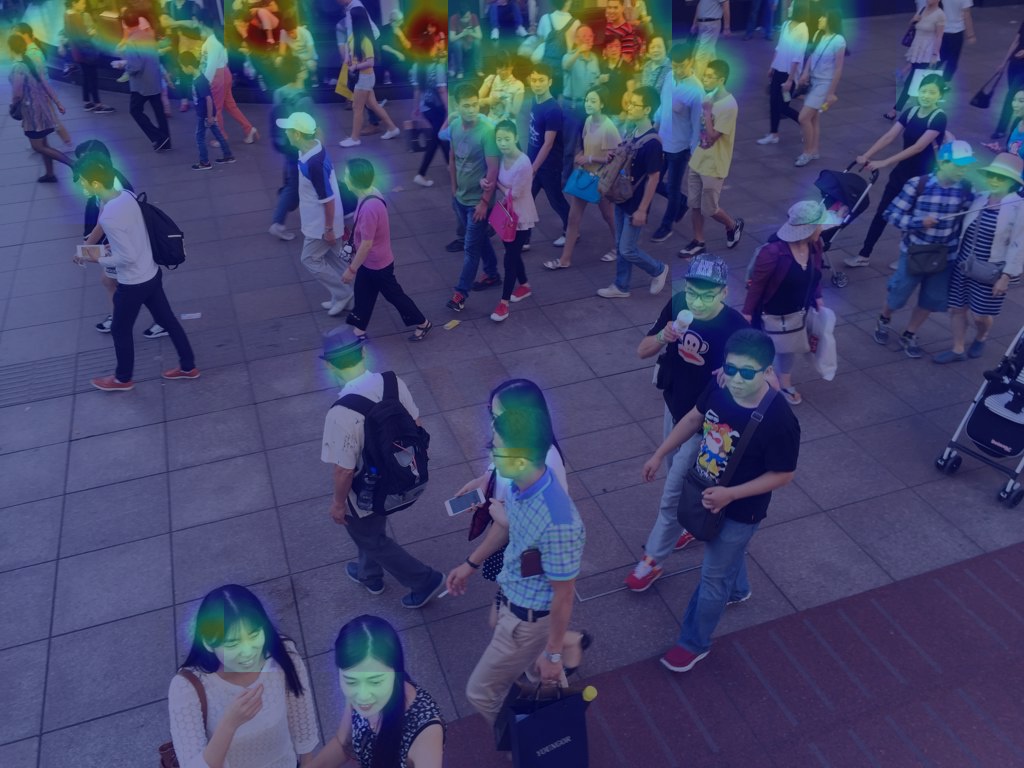}
\caption{ (Left) Saliency maps generated by the best GSP models on ShanghaiTech-A (top) and -B (bottom) datasets. (Right) Same for the best GAP models. GSP and GAP models exhibit similar activations, except GSP models are free from \textit{patchwise cancellation} effect due to single inference on full image.}
\label{fig:shanghaitech}
\end{figure}

\textbf{ShanghaiTech dataset: } This is the largest crowd counting dataset in terms of the number of counts (Table \ref{tab:stat_dataset}). It comprises two parts -- part A images are randomly collected from the web and part B is acquired from the busy streets of Shanghai. Table \ref{tab:a-shanghaitech-all} and \ref{tab:b-shanghaitech-all} enlist the comparative performance of GSP and GAP models on these subsets. Table \ref{tab:shanghaitech} reports the comparison with state-of-the-art approaches.

Our GSP-224 (part A) and GSP-128 (part B) models outperform state-of-the-performance approaches. Note that, although GAP models apparently provide good accuracy, their per patch error is pretty high indicating a considerable amount of patchwise cancellation. This is also evident from Figure \ref{fig:shanghaitech}. In this figure, both the best performing GSP and GAP models show similar saliency maps validating our claim that patchwise cancellation is heavily responsible for the comparatively poor performance of GAP models.

\begin{figure}[t]
\centering
\includegraphics[angle=90,scale=0.265]{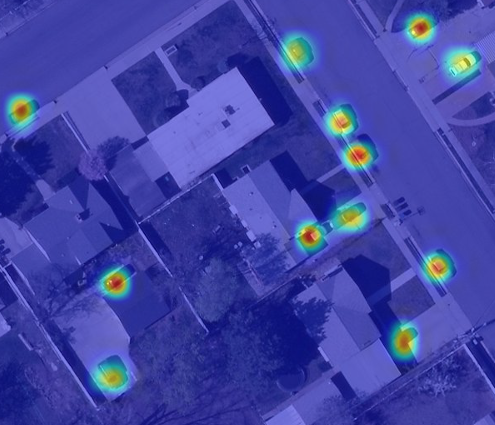}
\includegraphics[angle=90,scale=0.265]{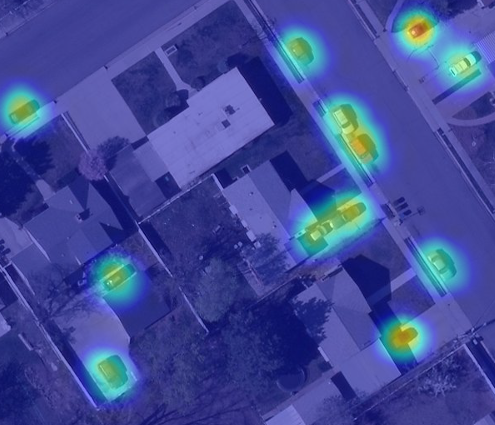}
\caption{Superimposed activation maps for GSP-64 (left) and GSP-224 (right) on the cropped image of COWC dataset. Activations are better localized the GSP-64 model.}
\label{fig:cowc}
\end{figure}

\begin{table}[]
\centering
\caption{GSP-GAP comparison on COWC dataset}
\label{tab:cowc-all}
\begin{adjustbox}{width=0.48\textwidth,center}
\begin{tabular}{c|l|ccccc}
\hline
\multicolumn{1}{l|}{Input} & Type & \multicolumn{1}{c}{MAE} & \multicolumn{1}{c}{RMSE} & \multicolumn{1}{c}{\%MAE} & \multicolumn{1}{c}{\%RMSE} & \%RMAE \\ \hline
\multirow{2}{*}{224} & GAP-C & 17.54 & 22.98 & 36.47 & 50.05 & 10.15 \\ \cline{2-7}
 & GSP & 8.85 & 13.01 & 10.70 & 14.99 & 5.12 \\ \hline
\multirow{2}{*}{128} & GAP-C & 20.05 & 43.18 & 13.89 & 17.72 & 11.60 \\ \cline{2-7}
 & GSP & 8.45 & 13.09 & 12.22 & 17.84 & 4.89 \\ \hline
\multirow{2}{*}{96} & GAP-C & 15.45 & 25.64 & 10.44 & 11.99 & 8.94 \\ \cline{2-7}
 & GSP & \textbf{8.20} & \textbf{12.53} & 11.13 & 16.38 & \textbf{4.75} \\ \hline
\multirow{2}{*}{64} & GAP-C & 24.34 & 45.16 & 19.30 & 24.09 & 14.09 \\ \cline{2-7}
 & GSP & 11.15 & 23.61 & \textbf{5.72} & \textbf{8.43} & 6.45 \\ \hline
\end{tabular}
\end{adjustbox}
\end{table}

\begin{table}[]
\centering
\caption{Results on COWC dataset}
\label{tab:cowc}
\begin{adjustbox}{width=0.35\textwidth,center}
\begin{tabular}{lcc}
\hline
Method & \%MAE & \%RMSE  \\ \hline
ResCeption \cite{cowc} & 5.78 & 8.09 \\ \hline
ResCeption taller 03 \cite{cowc} & 6.14 & 7.57 \\ \hline 
Ours (GSP-64) & \textbf{5.72} & 8.43 \\ \hline 
\end{tabular}
\end{adjustbox}
\end{table}

\textbf{COWC dataset: }
COWC contains very few training images (32) and the image sizes vary substantially (2220~$\times$~2220 to 18400~$\times$~18075), therefore it is an ideal test case for the main features of GSP.
Unlike the parking lot datasets, the COWC dataset contains images covering highways and residential areas and therefore cars in these images often appear to be entirely isolated objects in the roads or highways or parked in the residential streets. Each pixel covers 15 cm, resulting in the resolution of the cars ranging from 24 to 48 pixels. Because of the sparsity of the objects in the ultra-high-resolution training images, we extract $\sim$ 8000 samples of resolution 288~$\times$~288 centered on object sub-regions from the images prior to training. We do this to avoid training on a large number of negative samples that would be the case for random cropping.

For COWC, we could not provide per patch error metrics for GAP models in Table \ref{tab:cowc-all} since the test set contains only scalar counts as the ground truth. The smallest patch size (GSP-64) provides comparable performance to previously published results (Table \ref{tab:cowc}). We also see that the activations for GSP-64 are more concentrated on the objects in the image compared to that of GSP-224 (Figure \ref{fig:cowc}). This observation is consistent with our claim that the GSP models trained with smaller sample size tend to localize objects better, particularly when they are relatively isolated from each other.

\begin{table}[]
\centering
\caption{Results on Wheat-Spike dataset}
\label{tab:wheat-all}
\begin{adjustbox}{width=0.48\textwidth,center}
\begin{tabular}{c|l|ccccc}
\hline
\multicolumn{1}{l|}{Input} & Type & \multicolumn{1}{c}{MAE} & \multicolumn{1}{c}{RMSE} & \multicolumn{1}{c}{\%MAE} & \multicolumn{1}{c}{\%RMSE} & \%RMAE \\ \hline
\multirow{2}{*}{Full} & GAP & 132.25 & 153.77 & 13.82 & 16.40 & 13.24 \\ \cline{2-7}
 & GSP & 161.63 & 178.11 & 16.16 & 17.81 & 16.18 \\ \hline
\multirow{3}{*}{224} & GAP-C & 82.19 & 92.41 & 8.43 & 9.45 & 8.23 \\ \cline{2-7}
 & GAP-PS & 189.5 & 195.06 & 17.94 & 18.36 & 17.85 \\ \cline{2-7}
 & GSP & 108.19 & 134.01 & 10.37 & 12.38 & 10.83 \\ \hline
\multirow{3}{*}{128} & GAP-C & 91.00 & 106.07 & 9.45 & 11.07 & 9.11 \\ \cline{2-7}
 & GAP-PS & 279.75 & 284.67 & 24.93 & 25.08 & 24.92 \\ \cline{2-7}
 & GSP & 85.00 & 108.87 & 8.05 & 9.95 & 8.51 \\ \hline
\multirow{3}{*}{96} & GAP-C & \textbf{75.38} & \textbf{88.23} & 7.83 & 9.37 & 7.55 \\ \cline{2-7}
 & GAP-PS & 351.50 & 356.98 & 30.34 & 30.52 & 30.26 \\ \cline{2-7}
 & GSP & 80.00 & 100.63 & 7.94 & 9.93 & 8.01 \\ \hline
\multirow{3}{*}{64} & GAP-C & 87.50 & 99.02 & 9.19 & 10.46 & 8.76 \\ \cline{2-7}
 & GAP-PS & 514.00 & 519.63 & 41.29 & 41.49 & 41.07 \\ \cline{2-7}
 & GSP & 111.38 & 130.07 & 11.23 & 13.01 & 11.15 \\ \hline
\end{tabular}
\end{adjustbox}
\end{table}

\begin{figure}[t]
\centering
\includegraphics[scale=0.125]{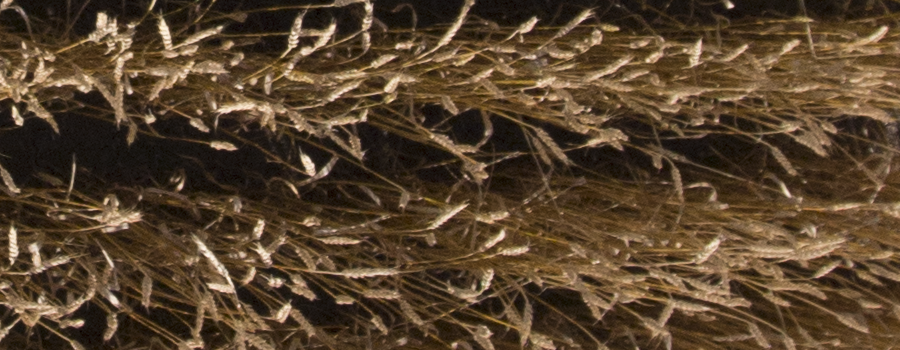} 
\includegraphics[scale=0.125]{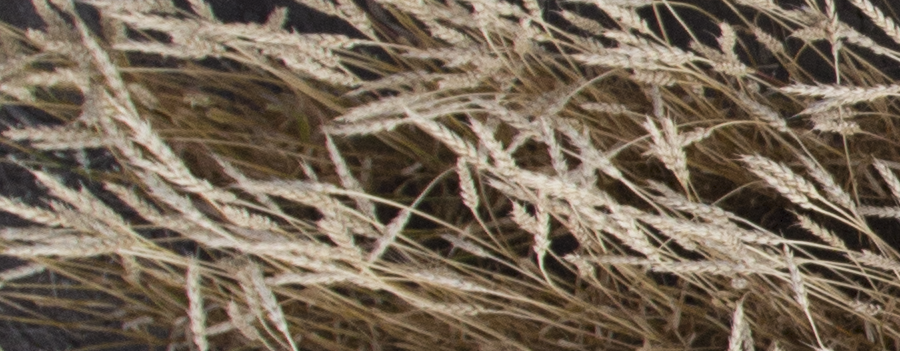} \\
\includegraphics[scale=0.125]{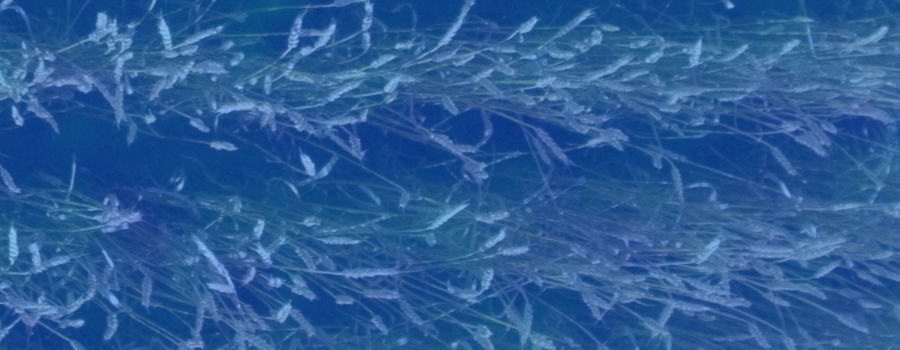}
\includegraphics[scale=0.125]{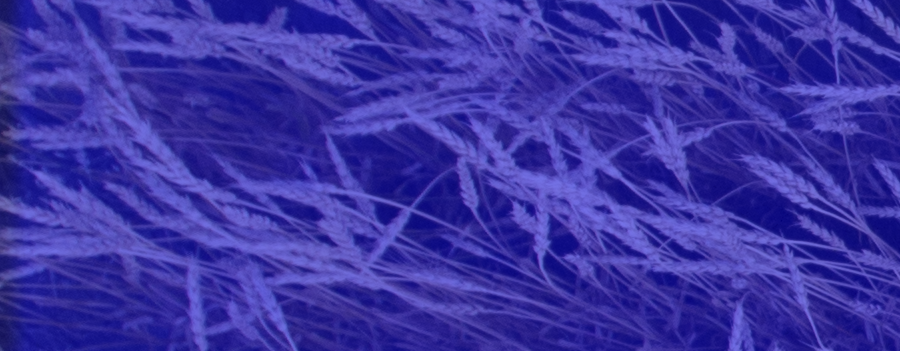} \\
\includegraphics[scale=0.125]{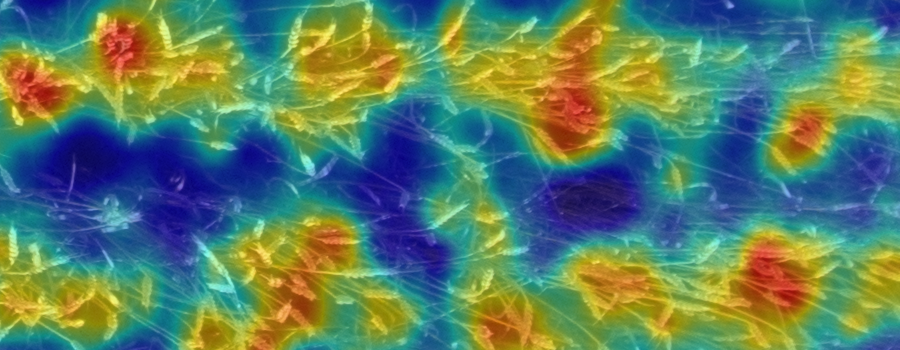}
\includegraphics[scale=0.125]{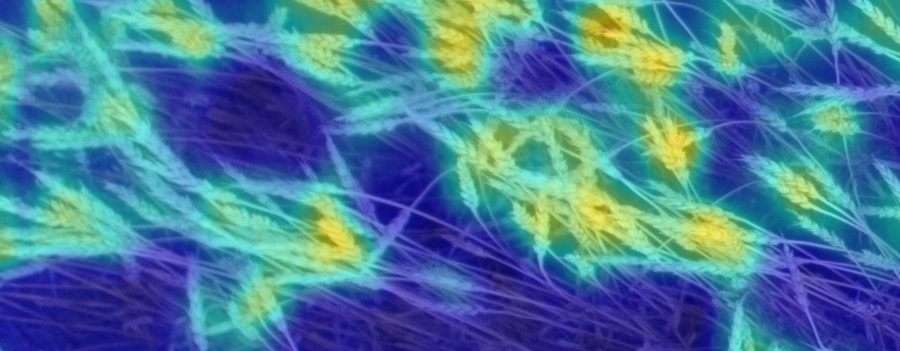}
\caption{Cropped sample images from Wheat-Spike dataset (top) with superimposed CAM generated by GSP-Full (middle) and GSP-96 (bottom) models. }
\label{fig:wheat}
\end{figure}

\textbf{Wheat-Spike dataset: }
This dataset \cite{wheat-spike} is a comparatively challenging one for object counting because of the irregular placement or collocation of wheat spikes. Out of 10 training samples, we use 8 for training and 2 for validation. Like COWC, the Wheat-Spike dataset is an ideal case study for GSP because of the low number of high-resolution training samples. Since the images are high-resolution and sub-regions inside a single image vary quite a bit in terms of brightness, perspective, and variable object shape resulting from natural morphology and wind motion, there are many features inside a single image that any suitable architecture should exploit without memorization or overfitting.

Table \ref{tab:wheat-all} reports the comparative performance of GSP and GAP models on this dataset. Although the summed up count for GAP seems to be more accurate than the corresponding GSP models, the surprisingly high aggregate of per-patch error again explains the effect of patchwise cancellation of under/overestimates with GAP models.

Also, the error for the GSP model trained with full-resolution images is quite high -- 161.63, about 16\% \textit{MAE} compared to the average count of 1000. GSP-96 provides the best performance with MAE of 80.00 (8\% of average count). Figure \ref{fig:wheat} shows cropped samples, their superimposed activation maps from GSP-Full model (middle), and GSP-96 (right). The GSP-96 model is able to identify salient regions in the image well, but for GSP-Full models, it tries to blindly memorize the count from only eight high-resolution images, which is clearly evident from the very uniform heatmap distribution all over the image regardless of foreground and background.

\subsubsection*{Acknowledgments}
This research was undertaken thanks in part to funding from the Canada First Research Excellence Fund and the Natural Sciences and Engineering Research Council (NSERC) of Canada.

\section{Conclusion and Future work}

In this paper, we introduce the global sum pooling operation as a way to train one-look counting models without overfitting on datasets containing few high-resolution images. With detailed experimental results on several datasets, we show that our GSP model, trained with small numbers of input samples, provides more accurate counting results than existing approaches. Also, when the GSP model is trained on small patches, it indirectly receives a weak supervision regarding object position and learns to localize objects better. This is also true for a GAP model, but it lacks the capability to infer counts from full-resolution test images and suffers from high per-patch errors for patchwise inference. This makes GAP unreliable for object counting on variable-sized, high-resolution images. Although we have only addressed object counting in this study, we believe that GSP could be applied to other computer vision tasks, such as classification or object detection. For these tasks, we expect that the scaling property of GSP may be able to utilize the features of image sub-regions over multiple spatial scales better than models that employ GAP or FC layers. In that case, the requirement for fixed-resolution images for many object detection or classification models can be eliminated. We plan to investigate these directions as future work.


{\small
\bibliographystyle{ieee}
\bibliography{egbib}
}

\end{document}